\documentclass[conference]{IEEEtran}
\IEEEoverridecommandlockouts

\usepackage{graphicx}
\usepackage[bookmarks=false]{hyperref}
\usepackage{verbatim}
\usepackage{balance}

\usepackage{todonotes}
\usepackage[]{enumerate}

\usepackage[utf8]{inputenc}

\usepackage{amssymb} 
\usepackage[ruled,noend,linesnumbered]{algorithm2e}
\usepackage{amsmath} 
\usepackage{blindtext}
\usepackage{graphicx}

\usepackage{booktabs}

\usepackage{subcaption} 
\usepackage{caption}

\usepackage{tikz}
\usetikzlibrary{trees}
\usepackage{cite}
\usepackage{amsmath,amssymb,amsfonts}
\usepackage{algorithmic}
\usepackage{graphicx}
\usepackage{textcomp}
\usepackage{xcolor}
\def\BibTeX{{\rm B\kern-.05em{\sc i\kern-.025em b}\kern-.08em
    T\kern-.1667em\lower.7ex\hbox{E}\kern-.125emX}}

\newtheorem{theorem}{Theorem}

\newtheorem{problem}{Problem}
\newtheorem{definition}{Definition}
\newtheorem{proof}{Proof}

\newcommand{\NP}{\ensuremath{\mathbf{NP}}\xspace}
\newcommand{\RSF}{\texttt{RSF}}
\newcommand{\NN}{\texttt{NN(1)}}

\newcommand{\IRT}{\ensuremath{\tau_{\mathit{IRT}}}\xspace}
\newcommand{\RT}{\ensuremath{\tau_{\mathit{RT}}}\xspace}
\newcommand{\NNT}{\ensuremath{\tau_{\mathit{NN}}}\xspace}
\newcommand{\compactness}{\ensuremath{\mathit{compact}}\xspace}
\newcommand{\plusclass}{\ensuremath{\text{`{\tt +}'}}\xspace}
\newcommand{\minusclass}{\ensuremath{\text{`{\tt -}'}}\xspace}
\newcommand{\dist}{\ensuremath{d}\xspace}
\newcommand{\subdist}{\ensuremath{d_s}\xspace}
\newcommand{\euclidean}{\ensuremath{d_{E}}\xspace}
\newcommand{\cost}{\ensuremath{c}\xspace}
\newcommand{\true}{\ensuremath{\mathrm{true}}\xspace}
\newcommand{\false}{\ensuremath{\mathrm{false}}\xspace}
\newcommand{\idx}{\ensuremath{\mathit{idx}}\xspace}

\usepackage{amsmath}

\DeclareMathOperator*{\argmin}{arg\,min}

\begin{document}

\title{Explainable time series tweaking via irreversible and
  reversible temporal transformations\\
}

\author{
\IEEEauthorblockA{Isak Karlsson, Jonathan Rebane, Panagiotis Papapetrou \\
\textit{Department of Computer and Systems Sciences} \\
\textit{Stockholm University}, Stockholm, Sweden \\
\{isak-kar,jonathan,panagiotis\}@dsv.su.se} \and
\IEEEauthorblockA{Aristides Gionis \\
\textit{Department of Computer Science} \\
\textit{Aalto University}, Espoo, Finland \\
aristides.gionis@aalto.fi} 
}

\maketitle

\begin{abstract}
Time series classification has received great attention over the past decade with a wide range of methods focusing on predictive performance by exploiting various types of temporal features. Nonetheless, little emphasis has been placed on interpretability and explainability. In this paper, we formulate the novel problem of explainable time series tweaking, where, given a time series and an opaque classifier that provides a particular classification decision for the time series, we want to find the minimum number of changes to be performed to the given time series so that the classifier changes its decision to another class. We show that the problem is \NP-hard, and focus on two instantiations of the problem, which we refer to as reversible and irreversible time series tweaking. The classifier under investigation is the random shapelet forest classifier. Moreover, we propose two algorithmic solutions for the two problems along with simple optimizations, as well as a baseline solution using the nearest neighbor classifier.  An extensive experimental evaluation on a variety of real datasets demonstrates the usefulness and effectiveness of our problem formulation and solutions. 
\end{abstract}


\section{Introduction}
\label{sec:intro}
Time series classification has been the center of attention in the time series community for more than a decade. The problem typically refers to the task of inferring a model from a collection of labeled time series, which can be used to predict the class label of a new time series.\footnote{To appear in
    International Conference on Data Mining, 2018}
Example applications of time series classification include historical document or projectile point classification \cite{ye2011time},  classification of electrocardiograms (ECGs) \cite{kampouraki2009heartbeat}, or anomaly detection in streaming data \cite{rebbapragada2009finding}.

Several time series classification models have been proposed in the literature, including distance-based classifiers (see Ding et al. \cite{ding2008querying} for a thorough review), shapelet-based classifiers~\cite{ye2011time,ye2011time} along with optimizations for shapelet selection or generation~\cite{hills2014classification,Grabocka:2014es,WistubaGS15}, and ensemble-based classifiers~\cite{COTE}. Recently, the random shapelet forest (RSF) \cite{KarlssonPB16} has been proposed for classifying univariate and multi-variate time series. The main idea is to build a set of decision trees, where each feature corresponds to a \emph{shapelet}. The decision condition on an internal node is the presence or absence of a shapelet in a test time series example.

Despite its competitive performance in terms of classification accuracy on a large collection of time series datasets, RSF is an opaque classification model. It is, hence, not feasible to come up with any reasoning behind the predictions that could possibly be helpful to domain experts and practitioners. \emph{Interpretability} studies within the time series domain have been largely dominated by the explanatory power provided by shapelets, which are class-discriminatory subsequences extracted from training examples \cite{ye2011time,Lines:2012:STT:2339530.2339579,xing2011extracting}. However, a clear gap has been present within the time series domain regarding \emph{explainability}, which this study has sought to address.

Consider the task of binary time series classification, where a times series may belong to either the positive (\plusclass) or negative class (\minusclass). Our main objective in this paper is to study the following simple problem: given a time series $\mathcal{T}$ 
and an opaque classification model (e.g., an RSF) expressed by function $f(\cdot)$, such that $f(\mathcal{T}) = \minusclass$, we want to identify the minimum number of changes that need to be applied to $\mathcal{T}$ in order to switch the classifier's decision to the positive class. That is, we want to define a transformation of $\mathcal{T}$ to $\mathcal{T}'$, such that $f(\mathcal{T}') = \plusclass$. We call this problem \emph{explainable time series tweaking}. By solving this problem, practitioners will not only be able to understand the reasoning behind decisions produced by opaque time series classification models, but will also be able to take action to \emph{change} a given time series instance from an undesired state (e.g., \emph{sick}) to a desired state (e.g., \emph{healthy}).

\subsection{Examples}
We present two motivating examples for the problem of explainable time series tweaking.

\begin{figure}[t!]
    \centering
    \includegraphics[width=0.9\columnwidth]{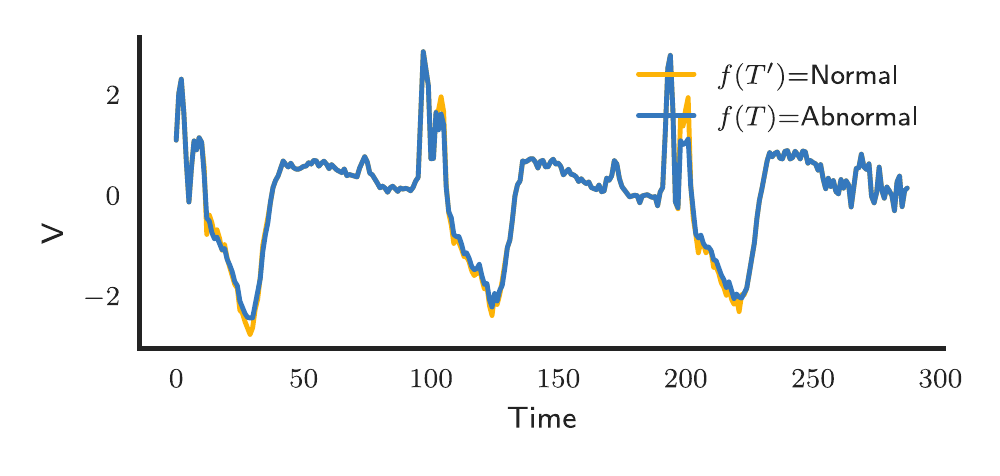}
    \caption{Abnormal vs.\  Normal heartbeat identification. The original time series is depicted in blue. We observe that a classifier $f$, classifies the input time series $\mathcal{T}$ as \emph{Abnormal} (blue curve). By applying time series tweaking, we change the classifier's decision to the normal class (yellow curve).
    }
    \label{fig:ecgexample}
\end{figure}

\smallskip
\noindent
\textbf{Example I: Abnormal vs.\ normal heartbeats.}
Consider an electrocardiogram (ECG) recording, such as the one shown in Figure~\ref{fig:ecgexample}.  The original signal (blue curve), denoted as $\mathcal{T}$, corresponds to a patient suffering from a potential myocardial disease. An explainable time series tweaking algorithm would suggest a transformation of the original time series to $\mathcal{T}'$ (yellow curve), such that the classifier considers it normal.

\begin{figure}[t!]
    \centering
    \includegraphics[width=0.9\columnwidth]{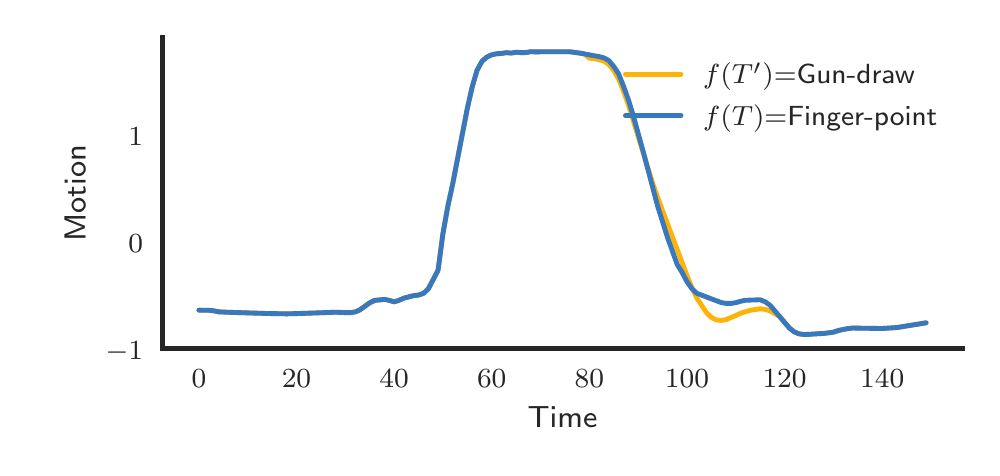}
    \caption{Gun-draw identification. The original time series is depicted in blue. We observe that a classifier $f$ classifies the input time series $\mathcal{T}$ as class \emph{Finger-point}. When transforming $\mathcal{T}$ to $\mathcal{T}'$ by changing two small segments (indicated in yellow) converts it to class \emph{Gun-draw}.
   }
    \label{fig:gunpoint}
\end{figure}

\smallskip
\noindent
\textbf{Example II: Gun-draw vs.\ finger-point.} 
Consider the problem of distinguishing between two motion trajectories, one corresponding to a gun-draw and the other to a finger-point. In Figure \ref{fig:gunpoint} we can see the trajectory of a regular finger pointing motion (blue time series), denoted as $\mathcal{T}$. The objective of explainable time series tweaking would be to suggest a transformation of $\mathcal{T}$ to $\mathcal{T}'$ (yellow curve), such that the classifier considers it a gun-point motion instead.

    
\subsection{List of contributions and organization} The main \textbf{contributions} of this paper are summarized as follows:
\begin{itemize}
    \item we formulate the novel problem of interpretable time series tweaking, and focus on two instantiations of the problem using the random shapelet forest classifier;
    \item we show that the problem is \NP-hard by a transformation from the Hitting Set problem;
    \item we propose two methods for solving the problem for the random shapelet classifier, which are based on \emph{shapelet feature tweaking}, along with optimization techniques;
    \item we provide an extensive experimental evaluation of the two proposed methods and compare them with a baseline Nearest Neighbour approach in terms of three metrics: cost, compactness and speed of transformation.
\end{itemize}

The remainder of this paper is organized as follows: in Section \ref{sec:related} we discuss the related work in the area of time series classification with emphasis on interpretability, while in Section \ref{sec:problem} we provide the formal problem formulation. In Section \ref{sec:methods} we describe the two proposed methods, along with optimization strategies and theoretical properties, while in Section \ref{sec:experiments} we present our experimental evaluation and results. Finally, in Section \ref{sec:conclusions} we conclude the paper and provide directions for future work. 

\section{Related Work}
\label{sec:related}
The majority of time series classification methods typically rely on instance-based classification techniques, For example, the $k$-Nearest Neighbor ($k$-NN) classifier, employs various similarity (or distance) measures, of which the most common and simplest is the Euclidean norm. To improve accuracy, elastic distance measures have been proposed, such as dynamic time warping (DTW) or longest common sub\-sequence \cite{lcss} and variants, e.g., cDTW \cite{sakoechiba}, EDR \cite{Chen05robust}, ERP \cite{Chen04onthe}, which are robust to misalignment and time warps. By regularization using, e.g., a band \cite{ratanamahatana2004everything}, the search performance and generalization behavior of $k$-NN can be greatly improved \cite{ding2008querying}. For a more complete overview of instance based univariate time series classifiers, the reader is referred to, e.g., Ding et al.~\cite{ding2008querying}.

A growing body of research is related to the domain of interpretable models, in which investigators have sought to provide greater clarity to decisions made by machine learning classifiers \cite{DBLP:journals/corr/RibeiroSG16,koh2017understanding, vellido2012making}. Such a need for interpretability often stems from a stakeholder desire to trust a model in order to find it useful; a trust which can be built both through the transparency of the model itself and post-hoc interpretability such as from local explanations \cite{DBLP:journals/corr/Lipton16a}.  As mentioned, a variety of studies in the time series domain highlight shapelets as the main vehicle for providing interpretability \cite{ye2011time,Lines:2012:STT:2339530.2339579,xing2011extracting} with at least one study providing an alternative Symbolic 
Aggregate approximation (SAX) combined with a vector space model approach \cite{senin2013sax}. 

Moreover, instance-based classifiers are supplemented by feature-based classifiers that typically use class-discriminant features, called shapelets \cite{ye2011time}, which correspond to time series subsequences with high utility, measured by different discriminative measures, such as information gain \cite{Shannon:1948iy}. For shapelet-based classifiers, the idea is to consider all sub\-sequences of the training data recursively in a divide-and-conquer manner, while assessing the quality of the shapelets using a \emph{scoring} function to estimate their discriminative power, constructing an interpretable \emph{shapelet tree} classifier \cite{ye2011time}. 

Shapelet transformation is one instance of a more general concept of feature generation, which has been thoroughly investigated  for time series classification. For example, the generated features can range from statistical features \cite{nanopoulos01feature, deng2013time} to interval-based features \cite{rodriguez2005support} or other interpretable features, such as correlation or entropy \cite{fulcher2014highly}. A typical grouping of features produced by these transformations includes: correlation-based, auto-correlation-based, and shape-based, each denoting similarity in time, change, and shape, respectively. For example, a time series forest based on interval features, such as averages, standard deviations and slope has been proposed by Deng et al.\ \cite{deng2013time} and a transformation based on time series bag-of-words Baydogan et al.\ \cite{Baydogan:2013ey}. Moreover, in order to achieve performance improvements, Hilles et al.  \cite{hills2014classification} introduce a heuristic approach for providing an estimation of the shapelet length. The described optimization algorithm repeatedly selects the ten best shapelets in a subset of ten randomly selected time series, searching for sub\-sequences of all possible lengths. 

Regarding multivariate time series classification methods, a shapelet forest approach has been introduced by Patri et al.\ for heterogeneous sensor data \cite{patri2014extracting}. The algorithm employs the Fast Shapelet selection approach for extracting the most informative shapelets per dimension. In a similar manner, a shapelet tree is built from each time series dimension \cite{cetin2015shapelet} using several additional techniques for providing search speedups. Moreover, various voting approaches are evaluated for providing the final classification label, demonstrating that one shapelet tree per dimension outperforms shapelets defined over multiple dimensions \cite{cetin2015shapelet}. More recently, the generalized random shapelet forest has been proposed for univariate and multivariate time series classification, by expanding the idea of random shapelet trees and randomly selecting shapelet features per dimension \cite{KarlssonPB16}. While this approach can achieve competitive performance against existing classifiers in terms of classification accuracy it is a black-box classifier with limited interpretability and explainability of the predictions.


Complementary to interpretability, a number of studies have focused on actionable knowledge extraction, where the focus is placed on identifying a transparent series of input feature changes intended to transform particular model predictions to a desired output with low cost. Many actionability studies exist with a business and marketing orientation, investigating actions necessary to alter customer behaviour for mostly tree-based models \cite{yang2007extracting,karim2013decision}. In addition, several studies place particular focus on actionability which can be performed in an efficient and optimal manner \cite{du2011efficient, turney1994cost}. For example, Cui et al. specified an algorithm to extract a knowledgeable action plan for additive tree ensemble models under a specified minimum cost for a given example \cite{Cui:2015:OAE:2783258.2783281}. Similarly, an actionability study by Tolomei et al. investigated actionable feature tweaking in regards to converting true negative instances into true positives; employing an algorithm which alters feature values of an example to the point that a global tree ensemble prediction is switched under particular global cost tolerance conditions~\cite{tolomei2017interpretable}.

Despite the expansion of explainability, this is an unexplored prospect within the time series domain. In this paper, we study the problem of altering the prediction of examples, through the alteration of examples themselves, such that the prediction of a tree ensemble is changed with minimal cost. Moreover, we achieve such class alterations in an effective and efficient manner, proving and addressing the \NP hard nature of the problem in accord with several optimization strategies. We then examine the real-world relevancy of this approach in regards to both medical and biomechanical time series datasets.

\section{Problem formulation}
\label{sec:problem}
In this section, we present our notation, and formally define the problem of explainable time series tweaking.

\begin{definition} 
\textbf{(Time series)}
A {time series} $\mathcal{T}=\{T_1,\ldots,$ $T_m\}$ is an ordered set of real values, sampled at equal time intervals, where each $T_i \in \mathbb{R}$.
\end{definition} 

In this paper, we only consider uni-variate time series, but the proposed framework and methods can be easily generalized to the multi-variate case. For the remainder of this paper, we will refer to uni-variate time series simply as time series.

A local, continuous segment of a time series is called a \emph{time series sub\-sequence}.

\begin{definition} \textbf{(Time series sub\-sequence or shapelet)} 
 Given a time series $\mathcal{T}$, a \emph{time series sub\-sequence or shapelet \cite{ye2011time}} of $\mathcal{T}$ is a sequence of $\ell$ contiguous elements of $\mathcal{T}$, denoted as $\mathcal{T}[s,\ell] = \{T_{s}, \ldots, T_{s+\ell-1}\}$, where $s$ is the starting position and $\ell$ is its length.
\end{definition}



Time series classification mainly relies on the chosen distance or similarity measure used to discriminate between instance pairs. The main task is to employ a distance function $\dist(\cdot)$ that compares two time series of equal length, and then given a time series sub\-sequence (corresponding to a candidate discriminant shapelet) identify the closest sub\-sequence match in the target time series. Depending on the application domain and the nature of the time series, various distance measures can be used. 

\begin{definition} \textbf{(Time series sub\-sequence distance)} 
Given two time series $\mathcal{S}$ and $\mathcal{T}$ of lengths $\ell$ and $m$, respectively, such that $\ell \leq m$, the \emph{time series sub\-sequence distance} between $\mathcal{S}$ and $\mathcal{T}$, is the minimum distance between $\mathcal{S}$ and any sub\-sequence of $\mathcal{T}$ of length $\ell$, i.e.:
\begin{equation}
  \subdist(\mathcal{S}, \mathcal{T}) = \min_{s=1}^{m-\ell+1} 
  \left\{ \dist (\mathcal{S}, \mathcal{T}[s,\ell]) \right\} \ .
\end{equation}
\end{definition}
A typical instantiation of $\dist(\cdot)$, given two time series $\mathcal{T}$ and $\mathcal{T'}$ of equal length $\ell$, is the Euclidean distance, i.e.:
\begin{equation}
 \dist(\mathcal{T},\mathcal{T}') = \euclidean (\mathcal{T},\mathcal{T}') =  \sqrt{ \sum_{i=1}^{\ell}(\mathcal{T}_{i}-\mathcal{T}_{i}')^2} \ .
\end{equation}

\begin{definition} \textbf{(Time series classification function)} %
  Given a time series $\mathcal{T}$ and a finite set of class labels $\mathcal{C}$, a classification function is a mapping
  $f$ from the set of all possible time series to the set $\mathcal{C}$, such that:
\[
f(\mathcal{T}) = \hat{y} \in \mathcal{C} \ .
\]
\end{definition}
Note that $\hat{y}$ denotes the predicted class for $\mathcal{T}$, and $f$ can be any type of time series classification function.

In this paper we study the problem of \emph{explainable time series tweaking}, which is formulated below.

\begin{problem} 
\label{prob:general}
\textbf{(Explainable time series tweaking)} Given a time series $\mathcal{T}$, a desired class $y'$, and a classifier $f$, such that $f(\mathcal{T}) = \hat{y}$, with $\hat{y} \neq y$, we want to find a transformation function $\tau$, such that $\mathcal{T}$ is transformed to $\mathcal{T}' = \tau(\mathcal{T})$, with $f(\mathcal{T}') = y'$, and $\cost(\mathcal{T},\mathcal{T}')$ is minimized, where $\cost(\mathcal{T},\mathcal{T}')$ defines the cost of the transformation. We call a transformation that changes the class \emph{successful} and the transformation that minimizes the cost the \emph{most successful} transformation.
\end{problem}

Any distance or similarity measure can be employed as a cost function. In this paper, we use the Euclidean distance, and consider two instantiations of Problem \ref{prob:general}, where $f$ is the random shapelet forest (RSF) classifier \cite{KarlssonPB16}.

\section{Explainable time series tweaking}
\label{sec:methods}
In this section, we first formulate the problem of explainable time series tweaking, then describe the shapelet transformation function, which is the building block of our solution, followed by the two algorithms to tackle the problem.
In addition, we present simple optimization strategies for both algorithms. Finally we prove that the problem we study is \NP-hard.

\subsection{Problem formulation}

In short, an RSF, denoted as $\mathcal{R} = \{F_1, \dots, F_{|\mathcal{R}|}\}$, is a shapelet tree ensemble of size $|\mathcal{R}|$, where each $F_j$ denotes a shapelet tree, constructed using a random sample of shapelet features \cite{ye2011time}. Each shapelet tree $F_j \in \mathcal{R}$ comprises a set of $t$ decision paths $\{P^{(y_1)j}_{1}, \ldots, P^{(y_t)j}_{t}\}$, where $y_i$ is the decision class of path $i$. 

Let $p_{ik}^j$ denote the $k^{th}$ non-leaf node in the $i^{th}$ path $P^{(y_i)j}_{i}$, such that 
\[
P^{(y_i)j}_{i} = \{p_{i1}^j, \ldots, p_{iu}^j\} \rightarrow y_i \ ,
\]
where $u$ is the length of path $P^{(y_i)j}_{i}$, i.e., $|P^{(y_i)j}_{i}|=u$ and each $p_{ik}^j$ is described by a tuple
\[
\langle \mathcal{S}^{j}_{k}, \theta^{j}_{k}, \delta^j_{ik}\rangle 
\]
defining a condition over shapelet $\mathcal{S}^{j}_{k}$ using a distance threshold $\theta^j_{k} \in \mathbb{R}$ and a comparison operator $\{\leq, >\}$, such that $\delta^j_{ik}$ equals $-1$ or $1$ if the comparison operator is $\leq$ or $>$, respectively.




\begin{definition} \textbf{(Condition test)}
Given a non-leaf node $p_{ik}^j = \langle\mathcal{S}^{j}_{k}, \theta_{k}^j, \delta_{ik}^j\rangle$ of shapelet tree $F_j$ and a time series $\mathcal{T}$, a condition test of path $i$ on non-leaf node $k$ is defined as:
\begin{equation}
\phi(\mathcal{T}, p_{ik}^j) = 
\begin{cases}
\true, & \text{if } (\subdist(\mathcal{S}_{k}^j, \mathcal{T}) - \theta^j_k) \delta_{ik}^j \leq 0\\
    \false,              & \text{otherwise}.
\end{cases}
\end{equation}

More concretely, $\phi(\cdot)$ returns \true if $\mathcal{T}$ fulfills the $k$-th condition of the $i$-th path of the $j$-th tree.

\end{definition}

\begin{figure}
    \centering
    \includegraphics[width=0.45\columnwidth]{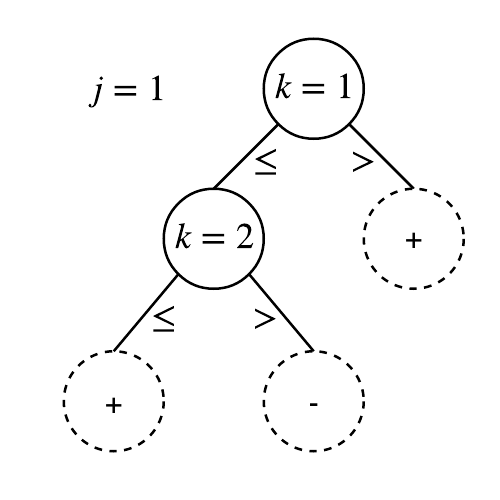}
    \caption{A simple decision tree example of two internal nodes and three leaf nodes.\label{fig:tree}}
\end{figure}

To clarify the notation consider the simple tree in Figure~\ref{fig:tree}, with two internal nodes and three terminal nodes. This tree can be converted into a set $F_1$ of 3 distinct paths: 
\begin{align*}
  P_{1}^{(\plusclass)1}&=\{\langle S^1_1, \theta^1_1, 1 \rangle\}  \\
  P_{2}^{(\minusclass)1}&=\{\langle S^1_1, \theta^1_1, -1\rangle, \langle S^1_2, \theta^1_2, 1\rangle\} \\ 
  P_{3}^{(\plusclass)1}&=\{\langle S^1_1, \theta^1_1, -1\rangle, \langle S^1_2, \theta^1_2, -1\rangle\}.
\end{align*}
Finally, observe that each non-leaf node performs a binary split depending on whether the time series sub\-sequence distance between $\mathcal{S}^{(i,k)}$ and $\mathcal{T}$ is within a distance range $\theta$. The decision label of $F_j$ for $\mathcal{T}$ is denoted as $y^j = f(\mathcal{T}, F_j)$, while the decision label of $\mathcal{R}$ for $\mathcal{T}$ is defined as $\hat{y} = f(\mathcal{T}, \mathcal{R}) = M(y^1,\ldots,y^{|\mathcal{R}|})$, where $M(\cdot)$ is the majority function. For more details on the actual structure and implementation of RSF the reader may refer to \cite{KarlssonPB16}.

The final step is to define a suitable transformation function $\tau(\cdot)$ for explainable time series tweaking. 
Given a time series example $\mathcal{T}$ and an RSF classifier $\mathcal{R}$, we define the transformation function $\tau(\cdot)$ used at each conversion step while traversing a decision path in each tree of the ensemble. Recall that our goal is to suggest the transformation of $\mathcal{T}$, such that the transformation cost is minimized and the classifier changes its classification decision. The smallest cost corresponds to the transformation that imposes the \emph{lowest Euclidean distance} between the original and transformed time series. 

We study two versions of $\tau(\cdot)$, hence defining the following two subproblems, \emph{reversible time series tweaking} and \emph{irreversible time series tweaking}.

\begin{problem} \textbf{(Reversible time series tweaking)}
\label{prob:re}
Given a time series $\mathcal{T}$, a desired class $y'$, and a RSF classifier $\mathcal{R}$, such that $f(\mathcal{T},\mathcal{R}) = \hat{y}$, with $\hat{y}\neq y'$, we want to transform $\mathcal{T}$ to $\mathcal{T}'=\tau(\mathcal{T})$, such that $f(\mathcal{T}', \mathcal{R}) = y'$, the Euclidean distance $\euclidean(\mathcal{T}, \mathcal{T}')$ is minimized, and $\tau(\mathcal{T})$ defines a sequence of transformations $\mathcal{T} \rightarrow \mathcal{T}^1 \rightarrow \mathcal{T}^2 \rightarrow \ldots  \rightarrow \mathcal{T}'$, where each subsequent transformation $\mathcal{T}^{i}$ \emph{\underline{can override}} any earlier transformation $\mathcal{T}^{j}$, with $j\leq i$.
\end{problem}

\begin{problem} \textbf{(Irreversible time series tweaking)}\label{prob:irr}
Given a time series $\mathcal{T}$, a desired class $y'$ and a RSF classifier $\mathcal{R}$, such that $f(\mathcal{T},\mathcal{R}) = \hat{y}$, with $\hat{y}\neq y'$, we want to transform $\mathcal{T}$ to $\mathcal{T}'=\tau(\mathcal{T})$, such that $f(\mathcal{T}', \mathcal{R}) = y'$, the Euclidean distance $\euclidean(\mathcal{T}, \mathcal{T}')$ is minimized, and $\tau(\mathcal{T})$ defines a sequence of transformations $\mathcal{T} \rightarrow \mathcal{T}^1 \rightarrow \mathcal{T}^2 \rightarrow \ldots  \rightarrow \mathcal{T}'$, where each subsequent transformation $\mathcal{T}^{i}$ \emph{\underline{cannot override}} any earlier transformation $\mathcal{T}^{j}$, with $j\leq i$.
\end{problem}

Note that Problem~\ref{prob:re} is a more general version of Problem~\ref{prob:irr} as the first one allows any change applied to the time series to be overridden by a later change, while the second one ``locks" the time series segments that have already been changed, hence not allowing for any change to be reversed. By restricting overriding transformations in Problem~\ref{prob:irr}, the Euclidean distance between the current and transformed time series is guaranteed to be monotonically increasing as more transformations are applied; hence allowing for \textbf{early abandoning} a transformation if the cumulative cost is above the currently best successful transformation. In contrast, reversible time series tweaking does not guarantee that the Euclidean cost is monotonically increasing, hence, it does not allow for early abandoning of the transformation. Despite this, we will show in Section~\ref{sec:rev} that a simple optimization can achieve substantial speedups for Problem~\ref{prob:re}.

\subsection{Time series tweaking}
Given a non-leaf node $p^j_{ik}$ containing a shapelet $\mathcal{S}^j_k$, and a threshold $\theta^j_k$, we define two types of time series tweaks: 
\begin{itemize}
\item \emph{increase distance}: if $\mathcal{S}^j_k$ exists in the current version of the time series (i.e., $\subdist(\mathcal{S}^j_k, \mathcal{T}) \leq \theta^j_k$) and the current $k^{th}$ condition demands that $\mathcal{S}^j_k$ does not (i.e., demanding that $\subdist(\mathcal{S}^j_k, \mathcal{T}) > \theta^j_k$), we want to increase the distance of all matches falling below $\theta^j_k$ to $> \theta^j_k$;
\item \emph{decrease distance}: if $\mathcal{S}^j_k$ does not exist in the current version of $\mathcal{T}$ (i.e., $\subdist(\mathcal{S}^j_k, \mathcal{T}) > \theta^j_k$) and the current $k^{th}$ condition demands that $\mathcal{S}^j_k$ does (i.e., it demands that $\subdist(\mathcal{S}^j_k, \mathcal{T}) \leq \theta^j_k$), we want to decrease the distance of its best match to $\leq \theta^j_k$.
\end{itemize}

\begin{figure}[t]
    \centering
    \includegraphics[width=0.6\columnwidth]{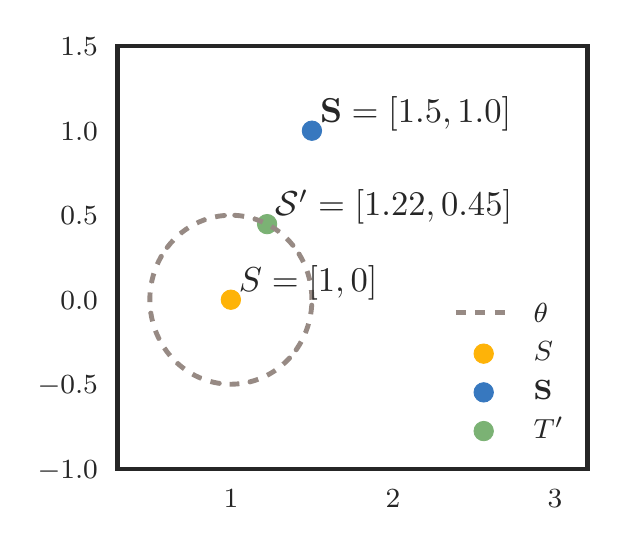}
    \caption{Example of moving the point $T$ to the closest point on the circle representing the distance threshold $\theta$, where the distance between $\dist(\mathcal{S}, \mathcal{T}) = \theta$.}
    \label{fig:move:example}
\end{figure}

As depicted in Figure~\ref{fig:move:example}, these time series tweaks can be achieved by considering any shapelet $\mathbf{S}$ as an $m$-dimensional point, and by defining an $m$-sphere with point $\mathcal{S}^j_k$ as its center and radius $\theta^j_k$. Intuitively, if $\euclidean(\mathbf{S},\mathcal{S}^j_k) \leq \theta^j_k$, then $\mathbf{S}$ falls inside the circle, and hence the resulting time series corresponds to the point on the circle that intersects the line connecting the two points. Given a desired distance threshold (radius) $\theta^j_k$, the transformed time series that has exactly the desired distance threshold is given by: 

\begin{equation}
 \tau_{\mathcal{S}}(\mathbf{S}, p^j_{ik}, \epsilon) = \mathcal{S}^j_k + \frac{\mathcal{S}^j_k - \mathbf{S}}{\lVert \mathcal{S}^j_k  - \mathbf{S} \rVert_2}(\theta^j_k+(\epsilon\delta^j_{ik}))
 \label{eq:move}
\end{equation}

\noindent where $\epsilon \in \mathbb{R}, \epsilon > 0$ is a parameter that control if the transformed time series distance fall inside the $m$-sphere ($\epsilon<0$), outside the $m$-sphere ($\epsilon > 0$) or exactly at the circumference ($\epsilon=0$). Note that in Equation~\ref{eq:move}, we use $\delta^j_{ik}$ to control the direction of the move, i.e., for condition $k$ with a $\leq$ test $\epsilon$ is negated and for conditions with a $>$ test $\epsilon$ is not. 

In summary, transforming a time series $\mathcal{T}$ predicted as $\hat{y}$ to a time series $\mathcal{T'}$ predicted as $y'$ for as single decision tree is a matter of changing the time series such that all conditions of the decision path, resulting in a transformation with the lowest cost, is successfully. Next, we will present two greedy algorithms for giving approximate solutions to Problem~\ref{prob:re} and Problem~\ref{prob:irr} using forests of randomized shapelet trees. 

\subsection{Greedy algorithm I: reversible tweaking}
\label{sec:rev}
Given an ensemble $\mathcal{R}$ of shapelets trees, where each tree $F_j$ is converted to a set of decision paths $\{P^{(y_1)}_{ij}, \ldots, P^{(y_t)}_{tj}\}$, a desired class label $y'$ and a transformation strength parameter $\epsilon$, which controls the amount of transformation applied, Algorithm~\ref{alg:rt} enumerate and apply the changes recommended by each condition $k$ for each path $i$ of all trees in the forest that is labeled with the desired label $y'$. 

\begin{algorithm}[t]
\caption{Reversible time series tweaking algorithm (\RT)\label{alg:rt}}
\SetEndCharOfAlgoLine{}
\SetKwInOut{Input}{input}\SetKwInOut{Output}{output}
\Input{A shapelet forest $\mathcal{R}$, a time series $\mathcal{T}$ and a desired class $y'$ and transformation strength $\epsilon$}
\Output{A transformed time series $\mathcal{T}'$}

 $\mathcal{T}' \leftarrow \mathcal{T}.copy$ \;
 $\text{c}_{min}\leftarrow \infty$ \;
 \For{$j\leftarrow 1$ \KwTo $|\mathcal{R}|$, $k\leftarrow 1$ \KwTo $|F_j|$} {
    \For{$i \leftarrow 1$ \KwTo $u$} {
        \If{$y^k = y' \land \phi(\mathcal{T}', p^j_{ik})$ is \false}{ \label{ln:check}
            $\mathbf{T} \leftarrow \mathcal{T}'.copy$ \;
            \eIf{$\subdist(\mathcal{S}_{k}^j, \mathcal{T}) \leq \theta^j_k$}{ \label{ln:check2}
                \While{$\subdist(\mathcal{S}_{k}^j, \mathcal{T}) \leq \theta^j_k$}{ \label{ln:while}
                    $\idx \leftarrow $ start index of sub\-sequence with lowest distance, $\subdist(\mathcal{S}^j_k, \mathbf{T})$  \;\label{ln:idx1}
                    $\mathcal{S}' \leftarrow \tau_\mathcal{S}(\mathbf{T}[\idx:\idx+|\mathcal{S}^j_k|], p^j_{ik}, \epsilon)$ \;
                    Assign $\mathcal{S}'$\ \KwTo $\mathbf{T}[\idx:\idx+|\mathcal{S}^j_k|]$ \; \label{ln:idx2}
                }
                \label{ln:last_if}
            } {
                $\idx \leftarrow $ start index of sub\-sequence with lowest distance, $\subdist(\mathcal{S}^j_k, \mathbf{T})$ \; \label{ln:idx_else}
                $\mathcal{S}' \leftarrow \tau_\mathcal{S}(\mathbf{T}[\idx:\idx+|\mathcal{S}^j_k|], p^j_{ik}, \epsilon)$ \;
                Assign $\mathcal{S}'$ \KwTo $\mathcal{T}[\idx:\idx+|\mathcal{S}^j_k|]$ \;
                \label{ln:last_else}
            }
        }
    }

    \If{$\cost(\mathbf{T}, \mathcal{T}) < c_{min} \land f(\mathbf{T}, \mathcal{R}) = y'$}{
        $\mathcal{T}' \leftarrow \mathbf{T}$ \;
        $c_{min} \leftarrow \dist(\mathcal{T}', \mathcal{T})$ \;
    }
}
 
 \Return{$\mathcal{T}'$}
\end{algorithm}

In Algorithm~\ref{alg:rt} transformations are applied one condition, from a path with the desired class, at a time. Consequently, the first step, on Line~\ref{ln:check}, is to check if the current condition test $k$ is fulfilled for the time series $\mathcal{T}'$, which we want to transform to $y'$. The check investigates if we need to apply any of the two tweaks in order for the current condition to hold. In the case where the condition does not hold, i.e., if $\phi(\cdot, \cdot)$ return \false, we check, on Line~\ref{ln:check2}, if there is a need to \emph{increase} or \emph{decrease} the distance to fulfill the $k^{th}$ condition. In the first case i.e., when the closest distance is larger than the threshold, but it needs to be smaller, the transformation is simple: we find the shapelet (starting at $\idx$ and ending at $\idx + |\mathcal{S}^j_k|$) with the closest distance and apply Equation~\ref{eq:move} to tweak the shapelet such that it distance is slightly smaller than $\theta^j_k$ and subsequently replaces the shapelet in $\mathbf{T}[\idx:\idx+|\mathcal{S}^j_k|]$ with the new sub\-sequence, $\mathcal{S}'$. In the second case, i.e., when the closest distance is smaller than the threshold but the distance needs to larger, the transformation is slightly more convoluted since there might exist many position where the distance is smaller than $\theta^j_k$. In the presented algorithm, we find and transform each lowest distance position incrementally, on Line~\ref{ln:while}, until there exists no sub\-sequence in the transformed time series with a distance smaller than $\theta^j_k$.

After all conditions $k,\ldots, u$ of the $i^{th}$ path has been applied, the algorithm computes the cost of transforming $\mathcal{T}$ to $\mathcal{T}'$, i.e., $\cost(\mathcal{T}, \mathcal{T}')$, and if this cost is lower than the \emph{best so far} \textbf{and} the classification according to $f(\mathbf{T}, \mathcal{R})$ has changed to $y'$, we record the current score as the lowest and keep track of the best transformation. This procedure is repeated for all paths, until the path with the lowest cost is returned.

\smallskip
\noindent
\textbf{Optimization via prediction ordering.}
Since the cost of prediction of the random shapelet forest is more costly than the cost of transforming a time series, one possible optimization is to compute all transformations, $\mathcal{T}'_1,\ldots, \mathcal{T}'_I$ for a particular time series $\mathcal{T}$ and order the transformed time series in increasing order according the transformation cost, $\cost(\mathcal{T}, \mathcal{T}'_i)$, where $i\in\{1,\ldots,K\}$. By ordering the prediction, the first transformation for which $f(\mathcal{T}', \mathcal{R}) = y'$ is \true, is by definition the transformation with lowest cost that also changes the class label. Although, this might seem a simple optimization, the pruning power and run\-time reduction is significant in practice, as seen in Section~\ref{sec:experiments}.

\subsection{Greedy algorithm II: irreversible tweaking}
The irreversible tweaking algorithm (\IRT{}), introduces a "locking" data structure that stores the start and end positions (i.e., $\idx$ and $\idx+|S^\cdot_\cdot|$ in Algorithm~\ref{alg:rt}) of transformed regions of the time series $\mathcal{T}$. As such, we modify Algorithm~\ref{alg:rt} to store these locked regions after transformations have been applied, at Line~\ref{ln:last_if} and Line~\ref{ln:last_else}. We also ensure that the sub\-sequence with the lowest distance does not overlap with a region has been previously \emph{locked}, by introducing an additional check on Line~\ref{ln:while} and after Line~\ref{ln:idx_else}. Note that by introducing the irreversible criterion, we are not guaranteed that the changes introduced by the algorithm changes the prediction of even the current tree $j$. However, as we show in Section~\ref{sec:experiments} this does not significantly affect the transformation cost, but the irreversible criterion allows the algorithm to produce more \emph{compact} transformations.

\smallskip
\noindent
\textbf{Optimization via early abandoning.}
For the reversible tweaking problem, early abandoning of transformation is not possible; however, if we specifically "lock" regions, as in the \IRT{} algorithm, of the time series that have already been transformed by an earlier condition $p^\cdot{ik}$, the cost is guaranteed to be monotonically increasing as we progress with further transformations. As such, as soon as a transformation is successfully, i.e., $f(\mathcal{T}', \mathcal{R}) = y'$, on Line~\ref{ln:check2}, conditions for which the partial cost is greater than or equal to $\cost(\mathcal{T}, \mathcal{T}')$ can be safely ignored, by introducing a partial cost indicator and check if the the current cost is increased above this value after each transformation, i.e., after Line~\ref{ln:last_else}. Using this simple technique, we can eliminate both predictions and transformations.

\begin{figure}[t]
    \centering
    \includegraphics[width=0.9\columnwidth]{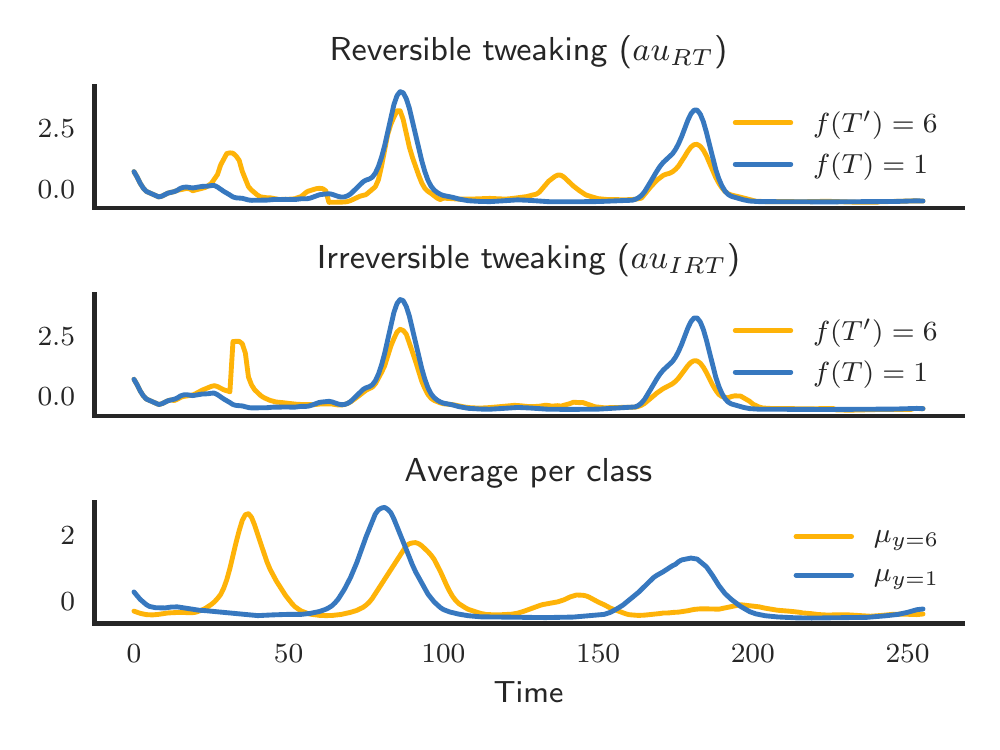}
    \caption{(top/middle) Example of transforming time series (blue) classified as $1$ by the classifier transformed to time series (yellow) labeled as $6$ by the classifier, using both the reversible and irreversible tweaking algorithms. (bottom) Average time series belonging to each of the classes, used for comparison. \label{fig:greedy:example:1}}
\end{figure}

An example of the both shapelet transformation algorithms are shown in Figure~\ref{fig:greedy:example:1} (top/middle), where a time series representing different insects flying through a audio recording device are transformed from being predicted as class $1$ (blue) and transformed to be predicted as $6$ (yellow). We can note that both tweaking algorithms increases the amplitude around time $t=35$ and reduces the amplitude around $t=100$ and $t=175$, all changes that seems to correspond well with the intuition provided by the average time series for each class (Figure~\ref{fig:greedy:example:1} (bottom)). 

\subsection{\NP-hardness}
Let us consider a very simple model where time series are sequences of binary values, and tree classifier test whether certain elements of the time series have a certain value. 
Let $\mathcal{T}$ be the time series, and let ${\mathcal{R}}=\{F_1,\ldots,F_m\}$ be the set of all decision trees in the ensemble.

\begin{theorem}
Given a time series $\mathcal{T}$ and an ensemble ${\mathcal{R}}$,  the problem of making the smallest number of changes in the time series so as to change the ensemble prediction is \NP-hard.
\end{theorem}

\begin{proof}
We consider the decision version of the problem, 
where a number $k$ is given and we ask whether there exists a solution
that requires at most $k$ changes in the time series.

We reduce a variant of the Hitting Set problem to the problem of 
explainable time-series tweaking (Problem~\ref{prob:general}).
An instance of the Hitting Set problem is the following: 
We are given a ground set $U$ of $n$ elements, 
subsets $S_1,\ldots,S_m \subseteq U$, 
and an integer $k$. 
We ask whether there is a set $H\subseteq U$
of cardinality $\lvert H\rvert$ at most $k$, 
so that $H\cap S_j \neq \emptyset$, for all $j=1,\ldots,m$, 
that is, whether there are at most $k$ elements in $U$
that ``hit'' all the sets $S_1,\ldots,S_m$.
Here we consider the variant where we ask whether 
there are at most $k$ elements in $U$ that hit 
{\em at least half} of the sets $S_1,\ldots,S_m$.
This is also an \NP-hard problem, as it is equivalent to Maximum Cover.

Given an instance of this variant of Hitting Set problem, 
we create an instance to the explainable time-series tweaking problem as follows.
We first create a time series of length $n$, 
where all its entries are 0s, that is, $\mathcal{T}[i]=0$, for all $i=1,\ldots,n$.
Then we create an ensemble ${\mathcal{R}}=\{F_1,\ldots,F_m\}$, 
so that there is a tree $F_j$ for each subset $S_j$.
In particular, the tree $F_j$ is constructed as follows.
If $i\in S_j$, then the tree $F_j$ contains a node
of the form 
``if $\mathcal{T}[i]=1$ then $\mathcal{T}$ is classified to class 1, 
otherwise $\langle${pointer to another node}$\rangle$''.
The tree $F_j$ is organized in an left-unbalanced manner, 
so that if none of these rules are satisfied,
they will all be checked.
The last (leftmost) leaf has the form
``if $\mathcal{T}[i]=1$ then $\mathcal{T}$ is classified to class 1, 
otherwise to class 0''.
It follows that the tree $F_j$ classifies the time-series to 1 
if and only if the series has a value equal to 1 in at least 
one position that corresponds to an element of the input set $S_j$.

We see that $\mathcal{T}$ is initially classified to class~0.
We ask whether it is possible to change at most $k$ positions
in $\mathcal{T}$ so that it is classified to class 1. One can easily see that the answer to this question is affirmative,
if and only if there exists a solution to the instance of the Hitting Set variant
that is given as input. 
Thus, we conclude that the explainable time series tweaking problem 
is \NP-hard.
\hfill\noindent$\Box$
\end{proof}

Note that we prove \NP-hardness for a very special case of our problem. 
As a result, the most general case of our problem, where we have real-valued time series, complex shapelets, 
and arbitrary decision trees in the ensemble is also \NP-hard.


\section{Experimental evaluation}
\label{sec:experiments}
\subsection{Experimental setup}
We evaluate the proposed algorithms on datasets from the UCR Time Series repository \cite{UCRArchive}. The datasets represent a wide range of different classification tasks varying from motion classification, e.g., \texttt{Gun Point} to sensor reading classification, e.g.,
\texttt{ECG200}. In the paper, we have selected all \textbf{binary classification tasks} to empirically evaluate the proposed time series tweaking algorithms. Hence, our task is to convert time series classified as $\hat{y}=1$ to $y'=-1$ and to convert time series classified as $\hat{y}=-1$ to time series classified as $y'=1$ by RSF; as such, the results presented in Table~\ref{tab:summary} are the \emph{average} of both transformations. Although, we limit the empirical evaluation to binary datasets, we note that the proposed algorithms can be used for multi-class problems. In the experiments, we set aside 20\% of the data for transformation and testing and use the remaining 80\% for training the model.

\subsubsection{Baseline}
The two proposed time series tweaking algorithms are compared to a baseline defined as the 1-nearest neighbour (1-NN) under the Euclidean distance, among the time series labeled as the target transformation label, which we call the training set; i.e., 
\begin{equation}
    \NNT(\mathcal{T}, y') = \argmin_{\{\mathcal{T}' | (\hat{y}, \mathcal{T}') \in \mathcal{D}, \hat{y}=y'\}} \euclidean(\mathcal{T}, \mathcal{T}').
\end{equation}
\noindent Note that the \NNT{} is guaranteed to find the transformation among the time series in the training set that minimizes the transformation cost as long as the transformation cost is the same as the 1-NN distance measure.

\subsubsection{Parameters}
The random shapelet algorithm requires several hyper-parameters to be set, namely the number of shapelets to sample at each node, the number of trees in the forest, and the minimum and maximum shapelet size. Since the purpose of this work is not to evaluate the effectiveness of the shapelet forest algorithm, the hyper-parameters are set to their default values, which amounts to $100$ random shapelets at each node and shapelets of all possible sizes. To have a viable number of paths to use for transformation, we let the learning algorithm grow $100$ trees. Moreover, we set the transformation strength for both the reversible and irreversible tweaking algorithms to $\epsilon=1$, which corresponds to relatively small changes.

\subsection{Performance metrics}
We compare the two algorithms and the baseline as the average cost over the test set, which we define as: 
\[
\cost_{\mu}(\tau, y') = \frac{1}{n}\sum_{i=1}^n \cost(\mathcal{T}_i, \tau(\mathcal{T}_i, y'))
\]
\noindent where $n$ is the number of time series in the test set not classified as $y'$. We report the average of $\cost_\mu(\cdot, y')$ with $y' \in \{\plusclass, \minusclass\}$. Moreover, we examine which fraction of the original time series must be altered under both the tweaking algorithms and the 1-NN approach.  Given $\mathcal{T}$, its transformation $\mathcal{T}'$, and a threshold $e \in \mathbb{R}$, assuming that $|\mathcal{T}|=|\mathcal{T}'|$ we define the compactness of a transformation of $\mathcal{T}$ to $\mathcal{T}'$ as
\[
     \compactness(\mathcal{T}, \mathcal{T}') = \frac{1}{|\mathcal{T}|}\sum_{i=1}^{|\mathcal{T}|} \mathit{diff}(T_i,T_i') \ , 
\]
\noindent where
\[
    \mathit{diff}(T_i,T_i') = \left\{
                \begin{array}{ll}
                  1 \text{, if } |T_{i}-T_{i}'| \leq e\\
                  0 \text{, otherwise.}
                \end{array}
              \right.
\]
\indent Note that a compactness of $1$ means that the entire time series is \emph{changed}, whereas a compactness of $0$ indicates that the transformed and original time series are identical. We report the average compactness, defined as:
\[
\compactness_\mu(\tau, y') = \frac{1}{n}\sum_{i=1}^n \compactness(\mathcal{T}_i, \tau(\mathcal{T}_i, y'))
\]
\noindent where $n$ is the number of time series in the test set not classified as $y'$. We report the average of $\compactness_\mu(\cdot, y')$ with $y' \in \{\plusclass, \minusclass\}$.

Finally, we examine the fraction of correct predictions, i.e., the accuracy, produced by our classifiers as a means of judging the trustworthiness of the classification approaches, and consequently the trustworthiness of the transformations.

\subsection{Results}

\begin{table*}[t!]
\centering
\caption{Summary of the results for the evaluation metrics. The best performing method for each metric is highlighted.\label{tab:summary}}

\begin{tabular}{l|rrr|rrr|rrr}
\toprule
                              & \multicolumn{3}{c|}{\textbf{Cost}} & \multicolumn{3}{c|}{\textbf{Compactness}} & \multicolumn{2}{c}{\textbf{Accuracy}}          \\
\textbf{Dataset}              & \RT                & \IRT              & \NNT      		    & \RT                & \IRT       		   & \NNT       & \RSF          & \NN{}          \\ \hline
BeetleFly                     & \textbf{7.3810}    & \textbf{7.3810}   & 26.6223   		    & \textbf{0.5737}    & \textbf{0.5737}    & 1.0000     & \textbf{0.8750}        & 0.7500       \\
BirdChicken                   & \textbf{4.5071}    & 4.5098            & 15.6695   		    & \textbf{0.5048}    & 0.5169              & 1.0000     & \textbf{1.0000}        & 0.6250       \\
Coffee                        & \textbf{1.1447}    & 1.1846            & 1.9178    		    & 0.3824     		 & \textbf{0.1809}     & 1.0000     & \textbf{1.0000}        & \textbf{1.0000}       \\
Computers                     & \textbf{2.2197}    & 2.5132            & 22.4809   		    & 0.4123     		 & \textbf{0.4044}     & 1.0000     & \textbf{0.7000}        & 0.4900       \\
DistalPhalanxOutlineCorrect   & \textbf{0.9314}    & 1.1150            & 1.1704    		    & 0.5917     		 & \textbf{0.4466}     & 0.9999     & \textbf{0.7886}        & 0.7143       \\
Earthquakes                   & \textbf{2.2725}    & 3.1455            & 30.0943   		    & \textbf{0.7449}    & 0.7577              & 1.0000     & \textbf{0.7826}        & 0.6630       \\
ECG200                        & \textbf{1.8730}    & 1.9080            & 4.1428    		    & 0.7976     		 & \textbf{0.7686}     & 1.0000     & \textbf{0.8750}        & 0.9500       \\
ECGFiveDays                   & \textbf{1.9722}    & 2.0158            & 4.2143    		    & 0.5215     		 & \textbf{0.4913}     & 1.0000     & \textbf{1.0000}        & 0.9944       \\
GunPoint                      & \textbf{1.9787}    & 1.9942            & 3.6975    		    & 0.4712     		 & \textbf{0.4460}     & 0.9998     & \textbf{1.0000}        & 0.9250       \\
Ham                           & \textbf{2.1744}    & 2.2187            & 7.8253    		    & 0.6791     		 & \textbf{0.6621}     & 0.9999     & \textbf{0.8605}        & 0.7907       \\
Herring                       & 1.2492             & \textbf{1.2488}   & 3.5817    		    & 0.4563     		 & \textbf{0.4060}     & 0.9999     & \textbf{0.5000}        & 0.3846       \\
ItalyPowerDemand              & \textbf{1.1791}    & 1.2645            & 1.3088    		    & 0.7262     		 & \textbf{0.6397}     & 0.9998     & \textbf{0.9726}        & 0.9589       \\
Lightning2                    & \textbf{3.2741}    & 3.9266            & 18.9703   		    & 0.7470     		 & \textbf{0.7071}     & 1.0000     & \textbf{0.6667}        & 0.6667       \\
MiddlePhalanxOutlineCorrect   & \textbf{0.6685}    & 0.9877            & 0.6791    		    & 0.6182     		 & \textbf{0.4493}     & 0.9999     & \textbf{0.8258}        & 0.7753       \\
MoteStrain                    & \textbf{2.4413}    & 2.5313            & 6.0249    		    & 0.5602     		 & \textbf{0.4834}     & 1.0000     & \textbf{0.9685}        & 0.9213       \\
PhalangesOutlinesCorrect      & \textbf{0.6979}    & 0.9568            & 0.7574    		    & 0.6186     		 & \textbf{0.5116}     & 0.9998     & \textbf{0.8421}        & 0.7782       \\
ProximalPhalanxOutlineCorrect & \textbf{0.5895}    & 1.0056            & 0.5326    		    & 0.6552     		 & \textbf{0.4121}     & 0.9997     & \textbf{0.8315}        & 0.8090       \\
SonyAIBORobotSurface1         & 1.7384             & \textbf{1.7260}   & 4.7213    		    & 0.4429     		 & \textbf{0.4394}     & 1.0000     & 0.9919        & \textbf{1.0000}       \\
SonyAIBORobotSurface2         & 1.8601             & \textbf{1.8566}   & 5.6126    		    & 0.4133     		 & \textbf{0.3584}     & 1.0000     & 0.9796        & \textbf{0.9949}       \\
Strawberry                    & \textbf{1.2082}    & 1.3628            & 1.2802    		    & 0.6644     		 & \textbf{0.5464}     & 0.9999     & 0.9695        & \textbf{0.9797}       \\
ToeSegmentation1              & \textbf{3.1200}    & 3.1436            & 14.7768   		    & 0.3871     		 & \textbf{0.3718}     & 1.0000     & \textbf{0.9259}        & 0.7407       \\
ToeSegmentation2              & \textbf{5.4407}    & 5.8238            & 17.8733   		    & 0.6173     		 & \textbf{0.5705}     & 1.0000     & \textbf{0.9697}        & 0.7879       \\
TwoLeadECG                    & \textbf{0.9112}    & 1.0671            & 1.3517    		    & 0.4966     		 & \textbf{0.4028}     & 0.9999     & \textbf{1.0000}        & 0.9957       \\
Wafer                         & \textbf{3.0135}    & 3.1419            & 8.6207    		    & 0.7152     		 & \textbf{0.6676}     & 0.9999     & 0.9958        & \textbf{0.9979}       \\
Wine                          & \textbf{0.5052}    & 0.9301            & 0.1708    		    & 0.7529     		 & \textbf{0.3452}     & 0.9996     & \textbf{1.0000}        & \textbf{1.0000}       \\
WormsTwoClass                 & \textbf{5.7723}    & 7.2023            & 28.7383   		    & 0.4416     		 & \textbf{0.4219}     & 1.0000     & \textbf{0.8269}        & 0.7308       \\ \hline
\textbf{Avg.}                 & \textbf{2.3132}    & \textbf{2.5329}   & \textbf{8.9552}    & \textbf{0.5733}     & \textbf{0.4942}     & \textbf{0.9999}     & \textbf{0.8924}        & \textbf{0.8240}  \\\bottomrule
\end{tabular}
\end{table*}

Tables~\ref{tab:summary} and~\ref{tab:runtime} show a comprehensive comparison in terms of the running time and the solution quality measured by the cost, transformation fraction, and run\-time per transformation. In Table~\ref{tab:summary} we observe, in regards to cost and compactness, that both the reversible tweaking \RT{} and irreversible tweaking \IRT{} approaches greatly outperform the nearest neighbor \NNT{} approach, with \RT{} demonstrating the best average cost by a small degree compared to \IRT{}, and \IRT{} showing the best level of compactness by a small degree compared to \RT{}. 

In terms of accuracy, \RSF{} on average provides more trustworthy predictions compared to \NNT{} and thus the explainable tweaking produced by \RSF{} would, not only result in less cost and more compactness, but potentially be considered more trustworthy by domain experts. In Table~\ref{tab:runtime} we present a run\-time comparison of \IRT{} against \RT{} with and without pruning. We observe that \RT{} with pruning provides the best run\-time performance on average, which can be explained by the fact that the relative cost of an \emph{ensemble prediction} is, on average, more costly than a \emph{transformation}. As such, the superior performance of optimized \RT{} can be attributed to its ability to prune more predictions than \IRT{}. In fact, for datasets with costly transformations (e.g., \texttt{PhalangesOutlinesCorrect}), the \IRT{} algorithm, which is able to prune 90\% of the transformations, outperform \RT{}. As a result, one should prefer \IRT{} when transformations are complex and the compactness of transformations are deemed important.

\begin{table*}[t!]
\centering
\caption{Summary of the run\-time of the two algorithms including the pruning power of the proposed optimization protocols. While \RT{} pruning does not prune any transformations, the \IRT{} pruning algorithm does. Hence, for \IRT{} the fraction of early abandoned transformations is the same as the fraction of pruned predictions.  \label{tab:runtime}}
\begin{tabular}{l|r|rrr|rr}
\toprule
\textbf{\textbf{Dataset}}     & $|\mathcal{T}|$ & \multicolumn{3}{c|}{\textbf{Run\-time} (seconds per transformation)} & \multicolumn{2}{c}{\textbf{Fraction of predictions pruned}} \\
                              &                 & \RT{} (no pruning)     & \RT{}               & \IRT{}             & \RT{}                       & \IRT{}             \\ \hline
BeetleFly                     & 512             & 1.763                  & \textbf{0.622}      & 0.646              & \textbf{0.181}              & 0.142              \\
BirdChicken                   & 512             & 1.966                  & \textbf{0.576}      & 0.622              & \textbf{0.328}              & 0.261              \\
Coffee                        & 286             & 1.099                  & \textbf{0.067}      & 0.092              & \textbf{0.774}              & 0.677              \\
Computers                     & 720             & 194.992                & \textbf{13.676}     & 29.358             & \textbf{0.902}              & 0.738              \\
DistalPhalanxOutlineCorrect   & 600             & 14.271                 & 0.918               & \textbf{0.772}     & \textbf{0.941}              & 0.863              \\
Earthquakes                   & 512             & 117.447                & \textbf{23.904}     & 46.285             & \textbf{0.724}              & 0.474              \\
ECG200                        & 96              & 2.194                  & \textbf{0.234}      & 0.269              & \textbf{0.719}              & 0.588              \\
ECGFiveDays                   & 136             & 1.173                  & \textbf{0.083}      & 0.104              & \textbf{0.718}              & 0.601              \\
GunPoint                      & 150             & 1.966                  & \textbf{0.144}      & 0.197              & \textbf{0.748}              & 0.598              \\
Ham                           & 431             & 17.898                 & \textbf{2.270}      & 3.494              & \textbf{0.761}              & 0.605              \\
Herring                       & 512             & 14.642                 & \textbf{1.710}      & 2.556              & \textbf{0.829}              & 0.725              \\
ItalyPowerDemand              & 24              & 3.217                  & 0.294               & \textbf{0.232}     & \textbf{0.917}              & 0.837              \\
Lightning2                    & 637             & 16.765                 & \textbf{3.867}      & 5.479              & \textbf{0.612}              & 0.390              \\
MiddlePhalanxOutlineCorrect   & 80              & 17.903                 & 1.404               & \textbf{1.158}     & \textbf{0.949}              & 0.906              \\
MoteStrain                    & 84              & 11.668                 & \textbf{1.062}      & 1.478              & \textbf{0.779}              & 0.457              \\
PhalangesOutlinesCorrect      & 80              & 69.298                 & 7.515               & \textbf{5.690}     & \textbf{0.949}              & 0.904              \\
ProximalPhalanxOutlineCorrect & 80              & 16.599                 & 1.351               & \textbf{1.160}     & \textbf{0.948}              & 0.881              \\
SonyAIBORobotSurface1         & 70              & 1.641                  & \textbf{0.111}      & 0.134              & \textbf{0.847}              & 0.692              \\
SonyAIBORobotSurface2         & 65              & 3.842                  & \textbf{0.373}      & 0.460              & \textbf{0.821}              & 0.592              \\
Strawberry                    & 235             & 22.119                 & \textbf{1.336}      & 1.727              & \textbf{0.923}              & 0.864              \\
ToeSegmentation1              & 277             & 3.208                  & \textbf{0.582}      & 0.755              & \textbf{0.600}              & 0.442              \\
ToeSegmentation2              & 343             & 2.574                  & \textbf{0.695}      & 0.790              & \textbf{0.350}              & 0.252              \\
TwoLeadECG                    & 82              & 2.400                  & 0.158               & \textbf{0.144}     & \textbf{0.941}              & 0.871              \\
Wafer                         & 152             & 18.709                 & \textbf{2.163}      & 34.402             & \textbf{0.815}              & 0.558              \\
Wine                          & 234             & 2.766                  & \textbf{0.147}      & 0.233              & \textbf{0.917}              & 0.788              \\
WormsTwoClass                 & 900             & 105.508                & \textbf{28.217}     & 40.302             & \textbf{0.566}              & 0.333              \\ \hline
\textbf{Avg.}                 &                 & 25.678                 & \textbf{3.595}      & 6.867              & \textbf{0.752}              & 0.617              \\ \bottomrule
\end{tabular}
\end{table*}

\subsection{Use-case examples}
In this section, we provide two use-case examples of the proposed time series tweaking framework by revisiting the motivating examples from Section~\ref{sec:intro}. 


\subsubsection{Electrocardiograms}
Revisiting the problem of heartbeat classification (Example I), we demonstrate a use-case example from the \texttt{ECG200} dataset, which contains measurements of cardiac electrical activity as recorded from electrodes at various locations on the body; each time series contains the measurements recorded by one electrode. The binary classification objective is to distinguish between \emph{Normal} and \emph{Abnormal} heartbeats.

\begin{figure}[t!]
    \centering
    \includegraphics[width=0.9\columnwidth]{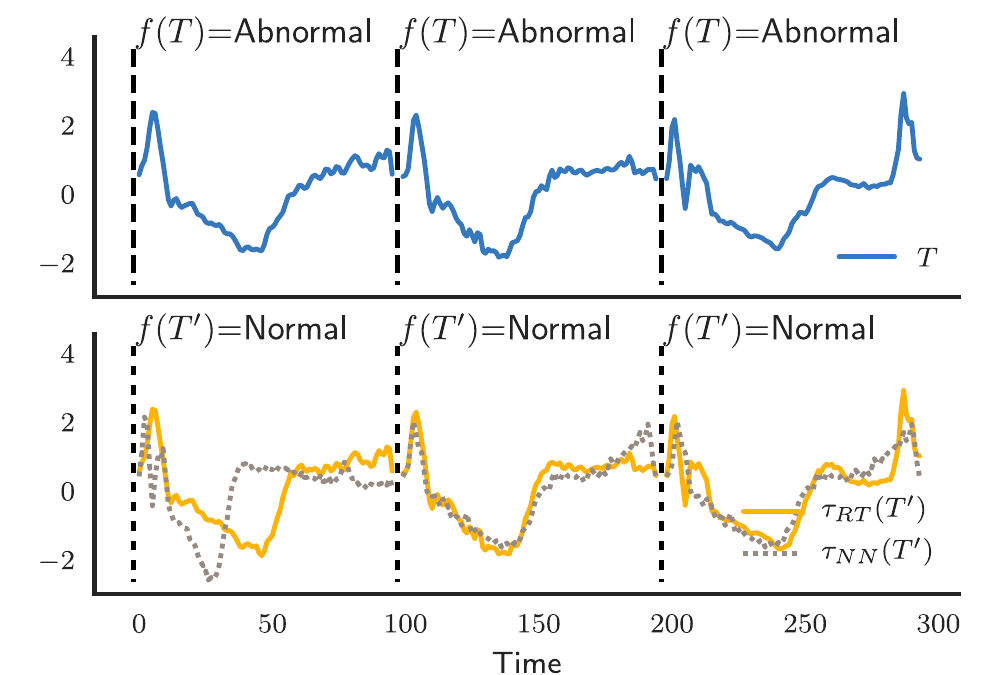}
    \caption{Abnormal vs.\ Normal heartbeat identification: the original time series is depicted in blue. We observe that a classifier $f$ labels the three segments of the input time series~$\mathcal{T}$ as \emph{Abnormal} (top). By applying \RT{}, we can transform these heartbeats to the normal class (bottom). We also show the transformations using $\NNT{}$ (bottom).} 
    
    \label{fig:ecg:case}
    \vspace{-0.5cm}
\end{figure}

In Figure~\ref{fig:ecg:case}, we observe that the original time series $\mathcal{T}$ (blue curve) exhibits a low-amplitude QRS complex, which may suggest a pericardial effusion or infiltrative myocardial disease \cite{cardio2015}, and is hence classified as \emph{Abnormal} by the \RSF{} classifier. Our explainable time series tweaking algorithm \RT{} suggests a transformation of the original time series to $\mathcal{T}'$ (yellow curve), such that the low-amplitude QRS complex is changed, by increasing the amplitude of the S-wave. This is illustrated by the yellow curve. Since \RT{} is the best performing transformation of the two proposed ones in terms of cost for this dataset, we apply it to $\mathcal{T}$ resulting in the classifier to label $\mathcal{T}'$ as \emph{Normal}. Moreover, we observe that the baseline competitor \NNT{} suggests a much costlier transformation (dotted curve).

\subsubsection{Gun-draw vs.\ finger-point}
Revisiting the problem of motion recognition (Example II), we demonstrate a use-case example from the \textit{Gun Point} dataset, which contains motion trajectories of an actor making a motion with his or her hand. The objective is to distinguish whether that motion corresponds to a gun-draw or to finger-pointing. 

\begin{figure}[t!]
    \centering
    \includegraphics[width=0.9\columnwidth]{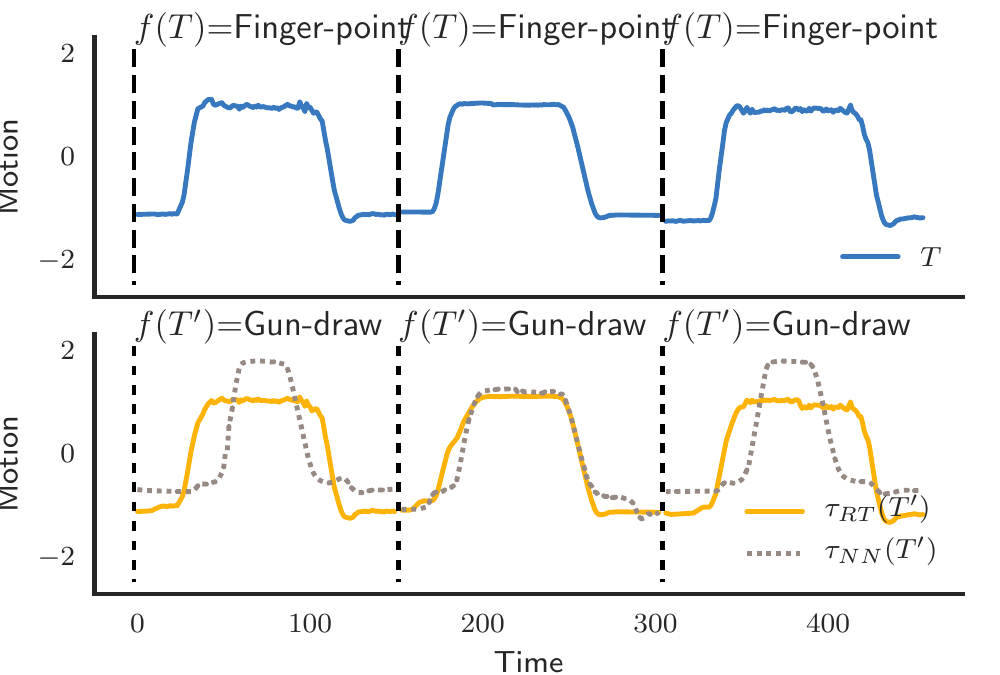}
    \caption{Gun-draw vs.\ Finger-point identification. The original time series is depicted in blue. We observe that \RSF{} classifies the three segments of the input time series $\mathcal{T}$ as \emph{Finger-point} (top). By applying \RT{}, we can transform these Finger-point motion trajectories to Gun-draw trajectories (bottom). We also show the transformations using $\NNT{}$ (bottom).} 
    \label{fig:gunpoint:case}
    \vspace{-0.5cm}
\end{figure}

In Figure \ref{fig:gunpoint:case} we observe the trajectory of a pointing motion (blue curve), for three consecutive motion recording, classified as \emph{Finger-point} by the \RSF{} classifier. By observing the bottom recording (yellow curve), we see the transformations needed to change the decision of \RSF{} to \emph{Gun-point}, using \RT{} (which has again achieved the lowest transformation cost, according to Table \ref{tab:summary}). Following our experimental findings, we also observe that the baseline competitor \NNT{} suggests a much costlier transformation (dotted curve).

\section{Conclusions}
\label{sec:conclusions}

In this study we have sought to exploit and expand upon the interpretability afforded by shapelets in the time series domain as a means of permitting explainability. We showed that the proposed problem formulation is \NP-hard and provided two instantiations of the problem using the random shapelet forest classifier. Experiments were performed to examine our approach in-depth and enable a comparison to a nearest neighbor solution in terms of Euclidean distance cost, compactness of transformations, and time needed for altering time series examples. We have demonstrated that the two proposed solutions outperform the baseline nearest neighbor solution in terms of cost and compactness, both of which are important factors in permitting actions pertaining to time series that are actually feasible in the sense that alterations can be realistically performed in a given domain. Future work includes the investigation of alternative distance measures, such as dynamic time warping, as well as expanding our approach to permit transformations exploiting trade-offs between cost and trustworthiness of classifier predictions.

\smallskip
\noindent
\emph{Reproducibility.}
Source code and data is available at: \newline \url{http://github.com/isakkarlsson/tsexplain}.

\section*{Acknowledgments}
This work was partly supported by grants provided by the Stockholm County Council and partly by the project Temporal Data Mining for Detective Adverse Events in Healthcare, ref.\ no.\ VR-2016-03372.
The work was also supported by 
three Academy of Finland projects  (286211, 313927, and 317085), 
and the EC H2020 RIA project ``SoBigData'' (654024). Finally, we acknowledge the UCR Time Series Classification Archive \cite{UCRArchive} for the datasets used in this paper.

\balance
\bibliographystyle{IEEEtran}
\bibliography{references}

\end{document}